\pdfoutput=1

\documentclass[11pt]{article}

\usepackage[]{ACL2023}

\usepackage{times}
\usepackage{latexsym}

\usepackage[T1]{fontenc}

\usepackage[utf8]{inputenc}

\usepackage{microtype}

\usepackage{inconsolata}

\usepackage{makecell} 

\usepackage{arydshln}
\usepackage{multirow}
\usepackage{comment}
\title{Data Augmentation for Low-Resource Keyphrase Generation}

\author{Krishna Garg \mbox{     }\mbox{     } Jishnu Ray Chowdhury \mbox{     }\mbox{     } Cornelia Caragea \\
        Computer Science \\ University of Illinois Chicago \\
        {\color{blue}\texttt{kgarg8@uic.edu} \mbox{     }\mbox{     }\mbox{     } \texttt{jraych2@uic.edu} \mbox{     }\mbox{     }\mbox{     } \texttt{cornelia@uic.edu}}
}

\begin{document}
\maketitle

\begin{abstract}
    Keyphrase generation is the task of summarizing the contents of any given article into a few salient phrases (or keyphrases). Existing works for the task mostly rely on large-scale annotated datasets, which are not easy to acquire. Very few works address the problem of keyphrase generation in low-resource settings, but they still rely on a lot of additional unlabeled data for pretraining and on automatic methods for pseudo-annotations. In this paper, we present data augmentation strategies specifically to address keyphrase generation in purely resource-constrained domains. We design techniques that use the full text of the articles to improve both present and absent keyphrase generation. We test our approach comprehensively on three datasets and show that the data augmentation strategies consistently improve the state-of-the-art performance. We release our source code at \url{https://github.com/kgarg8/kpgen-lowres-data-aug}.
\end{abstract}

\section{Introduction}
\label{sec:introduction}
Keyphrase generation (KG) helps in document understanding by summarizing the document in the form of a few salient phrases (or keyphrases). These keyphrases may or may not appear verbatim in the original text and accordingly, they are referred to as either \textit{present} or \textit{absent} keyphrases. The task has useful applications to many downstream tasks, e.g., document clustering \cite{hammouda2005corephrase}, matching reviewers to appropriate papers in the conference portals \cite{augenstein-etal-2017-semeval}, recommendation systems \cite{augenstein-etal-2017-semeval}, text classification \cite{wilson-etal-2005-recognizing, hulth-megyesi-2006-study, berend-2011-opinion}, index construction \cite{ritchie-etal-2006-find} and sentiment analysis and opinion mining \cite{wilson-etal-2005-recognizing, berend-2011-opinion}.

Prior works for keyphrase generation have largely focused on using large-scale annotated datasets for various domains - computer science (KP20k), news (KPTimes, JPTimes), webpages (OpenKP), etc. However, such large annotation datasets are not available in all domains (e.g., medicine, law, finance), either due to paucity in terms of available data or lack of domain expertise among the annotators or even the high annotation costs. This necessitates the focus on the low-resource domains.

The traditional ways to address low-resource keyphrase generation have been centered around using semi-supervised or unsupervised learning techniques \cite{ye-wang-2018-semi,wu-etal-2022-representation,chowdhury2021kpdrop}. 
For these methods, a lot of unlabeled data is necessary and needs to be curated for model training. 
The unlabeled data is further annotated automatically using keyphrase extraction methods and is used for pretraining the model or is used in the auxiliary task for multi-tasking. There are two limitations to these methods: (1) they have to still depend on additional large-scale unlabeled data, which may not be available always; and (2) the automatic annotation may not be accurate enough, especially when the off-the-shelf keyphrase generation or extraction models are pretrained on a different domain. 

In this paper, we develop data augmentation strategies for purely low-resource domains, which do not require acquiring unlabeled data for pretraining or automatic annotation approaches for unlabeled data (which may introduce errors). Inspired by \citet{garg-etal-2022-keyphrase} who showed the benefits of using information beyond the title and abstract for keyphrase generation, we leverage the full text of the documents (which is often ignored by prior works) and present ways for augmenting the text for improving both present and absent keyphrase generation performance.

\begin{table*}[t]
\centering
\def\columnseprulecolor{\color{grey}}
\resizebox{\textwidth}{!}{
  \begin{tabular}{c|p{47.5em}}
    \toprule
   \multicolumn{1}{c}{\rule{0pt}{2ex}\textbf{Methods}} &  \multicolumn{1}{c}{\textbf{Excerpts from different data augmentation methods}}\\[1ex] 
    \hline
    \rule{0pt}{3ex} \textsc{Title $||$ Abstract} & casesian : a \colorbox{myorange!50}{knowledge-based system} using statistical and experiential perspectives for improving the \colorbox{myorange!50}{knowledge sharing} in the \colorbox{myorange!50}{medical prescription} process [SEP] objectives : \colorbox{myorange!50}{knowledge sharing} is crucial for better patient care in the healthcare industry \\[0.5ex]
    \hline
     \rule{0pt}{3ex} \textsc{Aug\_TA\_SR} &  casesian : a knowledge based system using statistical and experiential perspectives for \colorbox{myblue!50}{better} the knowledge sharing in the medical \colorbox{myblue!50}{examination} prescription [SEP] objectives : knowledge sharing is crucial for \colorbox{myblue!50}{advantageously} patient \colorbox{myblue!50}{role} care in the healthcare industry\\[0.5ex]
    \hline
    \rule{0pt}{3ex}\textsc{Aug\_TA\_BT} & \colorbox{myblue!50}{cassian} : a knowledge-based system \colorbox{myblue!50}{that uses} statistical and experiential perspectives \colorbox{myblue!50}{to improve} the \colorbox{myblue!50}{sharing of} \colorbox{myblue!50}{knowledge} in the medical prescription process [SEP] objectives : knowledge sharing is \colorbox{myblue!50}{essential} to improve patient care in the \colorbox{myblue!50}{health} sector\\[0.5ex]
    \hline
    \rule{0pt}{3ex}\textsc{Aug\_TA\_KPD} & casesian : a \colorbox{myblue!50}{[MASK]} using statistical and experiential perspectives for improving the \colorbox{myblue!50}{[MASK]} in the \colorbox{myblue!50}{[MASK]} process [SEP] objectives : \colorbox{myblue!50}{[MASK]} is crucial for better patient care in the healthcare industry\\
    \hline
    \rule{0pt}{3ex}\textsc{Aug\_TA\_KPSR} & casesian : a \colorbox{myblue!50}{cognition based system} using statistical and experiential perspectives for improving the \colorbox{myblue!50}{noesis sharing} in the \colorbox{myblue!50}{checkup prescription} process [SEP] objectives : \colorbox{myblue!50}{noesis sharing} is crucial for better patient care in the healthcare industry\\
    \hline
    \rule{0pt}{3ex}\textsc{Aug\_Body} & \colorbox{myblue!50}{numerous methods have been investigated for improving the knowledge sharing process in medical prescription} \colorbox{myblue!50}{[SEP] case-based reasoning is one of the most prevalent knowledge extraction methods}\\
    \hline
    \rule{0pt}{3ex}\textsc{Gold Keyphrases} & \colorbox{myred!50}{case-based reasoning}, \colorbox{myorange!50}{medical prescription}, \colorbox{myorange!50}{knowledge-based system}, \colorbox{myorange!50}{knowledge sharing}, \colorbox{myred!50}{bayesian theorem}\\
    \bottomrule
  \end{tabular}
}
  \caption{An example depicting different augmentation methods used in the paper. The text is highlighted as follows: \colorbox{myblue!50}{\textsc{diversity} introduced in the augmented samples}, \colorbox{myred!50}{\textsc{absent keyphrases}}, \colorbox{myorange!50}{\textsc{present keyphrases}} (highlighted only in \textsc{Title || Abstract} for brevity). 
  Note that all \textsc{Aug} prefixed methods augment as a separate article to the original article \textsc{T || A}.  For specific details about each method, please refer to \S\ref{sec:methods}. Best viewed in color.}
\label{tab:cherryexample}
\end{table*}

Data augmentation in NLP has recently become a promising line of research to improve the state-of-the-art performance \cite{wei-zou-2019-eda, fadaee-etal-2017-data, li-caragea-2021-target, sun-etal-2020-mixup, xie2020unsupervised, feng-etal-2020-genaug, park-caragea-2022-data, yadav-caragea-2022-towards}. An ideal data augmentation technique is desirous to have the following characteristics: (1) to introduce diversity in training samples but neither too much (otherwise, training samples fail to represent the given domain) nor too less (otherwise, it leads to overfitting); (2) to be easy-to-implement; and (3) to improve model performance. 

Towards this end, we design and experiment with four data augmentation techniques (the first two being specifically designed for keyphrase generation) that remake the body\footnote{Body refers to the full text of the article excluding Title and Abstract.} of a given article and then augment it to the training data samples containing Title and Abstract (T $||$ A): (1) \textsc{Aug-Body-KPD} where the new training samples contain masked body (i.e., we drop present keyphrases with a certain probability from the body), (2) \textsc{Aug-Body-KPSR} where all the instances of present keyphrases (in contrast to random tokens as in the standard synonym replacement) in the body are replaced with their synonyms, (3) \textsc{Aug-Body-BT} where the body text is translated to an intermediate language and then back to the original language, (4) \textsc{Aug-Body-SR} where the standard synonym replacement is applied to random tokens of the body. In addition to augmentation with the body, we also provide methods for augmentation using T $||$ A. We depict the representative augmentation strategies in Table \ref{tab:cherryexample}. 

The intuition is that while augmenting the text if we further drop some of the present keyphrases, similar to Masked Language Modeling \cite{devlin-etal-2019-bert}, that makes the task harder and the model is forced to learn to generate the keyphrases. Introducing synonyms and back-translation further increases the diversity of the samples in a much controlled way. Recently, several full-text datasets have been proposed for the KG task, e.g., FullTextKP \cite{garg-etal-2022-keyphrase}, LDKP3K \cite{dl4srmahata2022ldkp}, and LDKP10K \cite{dl4srmahata2022ldkp}. We use two of these datasets, i.e., LDKP3K and LDKP10K, that contain scientific papers, along with a third dataset KPTimes \cite{gallina-etal-2019-kptimes} which mimics full-text keyphrase generation datasets but from a different domain, i.e., news. Through extensive experiments on the three datasets, we observe that although it is hard to improve the present keyphrase generation performance without sacrificing the absent keyphrase generation performance, our proposed augmentation approaches with the body consistently improve both. Moreover, the augmentation methods with body steadily surpass the performance of data augmentation methods that use only Title and Abstract.

In summary, the main contribution of the paper is to demonstrate data augmentation strategies for the keyphrase generation task particularly for purely low-resource domains (which have been under-explored). We present simple yet effective data augmentation methods using the full text of the articles and demonstrate large improvements over the state-of-the-art methods.

\section{Related Work}
\citet{meng-etal-2017-deep} first proposed to solve Keyphrase Generation as a sequence-to-sequence task using deep learning (encoder-decoder) methods. They proposed CopyRNN which uses the copy mechanism \cite{gu-etal-2016-incorporating} with the GRU-based encoder-decoder model. This was further extended by \citet{chen-etal-2018-keyphrase} to incorporate correlations between the predicted keyphrases (CorrRNN) and by \citet{yuan-etal-2020-one} to propose a mechanism to generate a sequence of a variable number of keyphrases (catSeq). Several other works approached the task using reinforcement learning \cite{chan-etal-2019-neural}, generative adversarial networks \cite{swaminathan-etal-2020-preliminary}, and hierarchical decoding \cite{chen-etal-2020-exclusive}. \citet{ye-etal-2021-one2set} further reframed the task as sequence-to-set generation instead of sequence-to-sequence generation and used the transformer model for the first time for this task. Later, \citet{garg-etal-2022-keyphrase, wu-etal-2022-representation, kulkarni-etal-2022-learning, wu-etal-2021-unikeyphrase} used other pretrained models like Longformer Encoder-Decoder, BART, KeyBART, and UniLM. In this paper, we constrain our focus to \textit{CatSeq} model \cite{yuan-etal-2020-one} and explore data augmentation strategies using CatSeq on three datasets. However, our augmentation strategies can be extended to work with other pre-trained models in future work.

\paragraph{Data Augmentation \& Keyphrase Generation.} 
Data augmentation has been explored in related tasks like Named-Entity Recognition \cite{dai-adel-2020-analysis, wang-henao-2021-unsupervised}, and Keyphrase Extraction \cite{veyseh2022improving, liu2018scientific}, but there have been minimal efforts for exploring data augmentation in Keyphrase Generation. Most of such works deal with augmentation of the candidate keyphrases (extracted using an off-the-shelf unsupervised keyphrase extraction method) to the ground truth keyphrases. \citet{ye-wang-2018-semi} generated synthetic ground truth labels for the additional unlabeled data. \citet{Shen_Wang_Meng_Shang_2022} generated silver labels in addition to the gold-labeled keyphrases using an automatic comparison and ranking mechanism. \citet{chen-etal-2019-integrated, santosh2021gazetteer} augmented keyphrases from semantically similar documents to improve keyphrase generation. 
In contrast, we deal mainly with the augmentation on the input side (i.e., augmenting text to the given articles instead of augmenting the ground-truth keyphrases). \citet{garg-etal-2022-keyphrase} used external information from various parts of the body and appended it to the T $||$ A of the given articles. Our data augmentation strategy is weakly inspired by this work and we use this work as one of the baselines for comparison. \citet{chowdhury2021kpdrop} proposed a data augmentation strategy similar to one of our augmentation methods (suffixed with \textsc{KPD}), i.e., randomly dropping {\em present} keyphrases from text.. We leverage the strategy further to drop the present keyphrases from even the body of the articles and then augment it to the articles themselves.

\paragraph{Low-Resource Keyphrase Generation.} \citet{wu-etal-2022-representation} presented a method for a low-resource setting where they utilized the major fraction of a large-scale dataset (KP20k) as unlabeled data for pretraining (using sophisticated pretraining objectives) and the smaller fraction of the dataset for fine-tuning. \citet{ye-wang-2018-semi} proposed a semi-supervised technique where they created synthetic keyphrases for the large-scale unlabeled data and also utilized the unlabeled data for training the model in 
a multi-tasking fashion. In contrast, our methods do not require acquiring any unlabeled data or pretraining or multi-task training but work with a few annotated samples. However, all the above works can very well complement our methods to further improve the performance.

\begin{table*}[t]
\small
\centering
\renewcommand{\arraystretch}{1}
\setlength\tabcolsep{8pt}
\begin{tabular}{lcccccccc} 
\hline
\Xhline{3\arrayrulewidth}
Datasets & \#Train & \#Dev & \#Test & Avg \#words & Avg \#kp & Avg kp-len & \% Present & \% Absent \\
\hline
LDKP3K$\spadesuit$
& 50,000 & 3,339 & 3,413 & 6,457 & 4.45 & 1.86 & 84.24 & 15.76 \\
LDKP10K$\spadesuit$
& 50,000 & 10,000 & 10,000 & 4,674 & 5.98 & 2.07 & 74.40 & 25.60 \\
KPTimes
& 259,923 & 10,000 & 20,000 & 948 & 4.03 & 2.17 & 48.44 & 51.56 \\
\hline
\Xhline{3\arrayrulewidth}
\end{tabular}
\caption{Statistics of the datasets. \textit{\#words}: number of words in the document, \textit{\#kp}: number of keyphrases, \textit{kp-len}: keyphrase length, \% Present (Absent): percentage of present (absent) keyphrases. $\spadesuit$ indicates \textit{medium} version of the original dataset. Note that the statistics are computed for the combined train, dev and test sets.}
\label{tab:datasets}
\end{table*}

\section{Methods}
In this section, we first describe the formulation of the keyphrase generation task. Next, we describe the baselines followed by the data augmentation strategies that we propose for keyphrase generation.

\paragraph{Problem Formulation.} 
Keyphrase Generation can be posited as a sequence-to-sequence generation task where the input is the text from a given article and the output is a sequence of keyphrases that summarize the article. Formally\footnote{We model the problem similar to \textsc{catSeq} as proposed by \citet{yuan-etal-2020-one}.}, the task can be denoted as follows:

Input: Title $||$ Sent$_1$ $||$ Sent$_2$ $|| ... ||$ Sent$_k$

Output: $kp_1$ $||$ $kp_2$ $|| ... || kp_n$

\noindent where $kp_i$ denotes a keyphrase, \textit{Sent$_j$} denotes a sentence from the abstract or from the body of the article, $||$ denotes any delimiter (e.g., [SEP] in this work).

\subsection{Baselines}

\noindent \textbf{\textsc{T $||$ A:}} This baseline contains all the training samples with Title and Abstract concatenated as T [SEP] A.

\noindent \textbf{\textsc{T $||$ A $||$ Body:}} For this baseline, we simply concatenate the body of the article to T $||$ A. This baseline was presented in the prior work by \citet{garg-etal-2022-keyphrase}.

\subsection{Data Augmentation Strategies}
\label{sec:methods}
Further, as discussed in \S\ref{sec:introduction}, we describe the data augmentation strategies created primarily using four ways of augmentation: dropout, synonym replacement (both keyphrase-specific and standard) and back-translation. We describe them as follows:

\vspace{1mm}
\noindent \textbf{\textsc{Aug\_Body:}} In this method, we augment the training set with the text from the body of each article, which doubles the total number of samples. That is, one sample is \textbf{\textsc{T $||$ A}} and the other is \textbf{\textsc{Body}} (i.e., sentences from the body of the article). 

\vspace{1mm}
\noindent \textbf{\textsc{Aug\_Body\_KPD:}} In this method, we first apply the dropout technique presented by \citet{chowdhury2021kpdrop} to the body of the article and then augment it (as above). The dropout technique is to mask some of the present keyphrases (particularly, all occurrences of a given keyphrase) in the body of the article.

\vspace{1mm}
\noindent \textbf{\textsc{Aug\_TA\_KPD:}} In this method of augmentation, we first apply the dropout technique to the T $||$ A, and then add it to the training set.

\vspace{1mm}
\noindent \textbf{\textsc{Aug\_Body\_KPSR:}} In this method, we replace all the present keyphrases in the body of the article with the corresponding synonyms from NLTK WordNet \cite{miller1995wordnet} and augment it to the training set. If a particular keyphrase does not have a corresponding synonym, we retain the original keyphrase. Notably, only a small number of keyphrases lack synonyms in the WordNet. For instance, we were able to find synonyms for 2936 (out of 3282) keyphrases for data augmentation on the Body, with 1000 samples of LDKP3K dataset. We show the statistics for the LDKP3K dataset in Table \ref{tab:statistics}.

\begin{table}[ht]
\small
\centering
\renewcommand{\arraystretch}{1.3}
\setlength\tabcolsep{6pt}
\begin{tabular}{lllll}
\hline
\Xhline{3\arrayrulewidth}
 & 1000 & 2000 & 4000 & 8000 \\
 \hline
Aug\_TA\_KPSR & 3386/ & 6705/ & 13374/ & 26757/ \\
& 3733 & 7385 & 14702 & 29398 \\
Aug\_Body\_KPSR & 2936/ & 5844/ & 11671/ & 23538/\\
& 3282 & 6515 & 13001 & 16171 \\
\hline
\Xhline{3\arrayrulewidth}
\end{tabular}
\caption{Statistics of the synonyms replaced/ total synonyms by \textsc{Aug\_Body\_KPSR} and \textsc{Aug\_TA\_KPSR} methods for LDKP3K dataset for four settings, i.e., 1000, 2000, 4000, 8000 samples.}
\label{tab:statistics}
\end{table}

\vspace{1mm}
\noindent \textbf{\textsc{Aug\_TA\_KPSR:}} This is similar to \textsc{Aug\_Body\_KPSR} but with the difference that we replace present keyphrases with their synonyms in the T $||$ A instead of the body of the article.

\vspace{1mm}
\noindent \textbf{\textsc{Aug\_Body\_BT:}} In this method, we backtranslate the body of the article from English to French and back to English using Opus-MT \cite{tiedemann-thottingal-2020-opus} pretrained translation models. The backtranslated (or equivalently, paraphrased) articles are then augmented as separate samples to the training set. During the translation of text from one language to another, we use temperature sampling with a temperature value equal to 0.7.

\vspace{1mm}
\noindent \textbf{\textsc{Aug\_TA\_BT:}} This method applies back translation model to the T $||$ A instead of the body and does augmentation similar to \textsc{Aug\_Body\_BT}.

\vspace{1mm}
\noindent \textbf{\textsc{Aug\_Body\_SR:}} We use the standard synonym replacement, i.e., we randomly select 10\% of the tokens from the body of a given article, replace them with their corresponding synonyms from NLTK Wordnet, and augment the text as a separate article to the training set.

\vspace{1mm}
\noindent \textbf{\textsc{Aug\_TA\_SR:}} We do augmentation similar to \textsc{Aug\_Body\_SR} but use the T $||$ A instead of body. 


\section{Experimental Setup}
\vspace{-2mm}
\subsection{Datasets}
We conduct experiments on three datasets for keyphrase generation. All these datasets contain the full text of the articles along with the keyphrase annotations. 1) \textbf{LDKP3K} \cite{dl4srmahata2022ldkp} contains computer science research articles from online digital libraries like ACM Digital Library, ScienceDirect and Wiley. It is a subset of KP20K corpus \cite{meng-etal-2017-deep} but each article now contains the full text instead of just the title and abstract. 2) \textbf{LDKP10K} \cite{dl4srmahata2022ldkp} expands a subset of articles from OAGkx dataset \cite{ccano2019keyphrase} to contain their full text. The articles are scientific publications curated from various domains. We use the \textit{medium} version of both LDKP datasets (each consists of 50,000 samples in the training set) to facilitate quality sampling of the articles for the low-resource setting while being mindful of the computational budget. 3) \textbf{KPTimes} \cite{gallina-etal-2019-kptimes} is a large-scale dataset with long news texts. To mimic KG datasets, we map the heading of the news article to \textit{Title}, and segment the main body of the news article into a maximum of 300-words\footnote{The length was chosen on a similar scale as the average length of abstracts in LDKP10K, which is about 260 words.} \textit{Abstract} and the rest of the text as \textit{Body}. We choose KPTimes to validate our observation on an altogether different domain. Datasets' statistics are shown in Table \ref{tab:datasets}. Dataset preprocessing steps are outlined in Appendix \S \ref{sec:implementation}.

\begin{table*}[ht]
\small
\centering
\renewcommand{\arraystretch}{0.97}
\setlength\tabcolsep{10pt}
\begin{tabular}{lcccccccc}
\hline
\Xhline{3\arrayrulewidth}
\multirow{2}{*}{\textbf{LDKP3K}} & \multicolumn{2}{c}{1,000} & \multicolumn{2}{c}{2,000} & \multicolumn{2}{c}{4,000} & \multicolumn{2}{c}{8,000} \\
 & \multicolumn{1}{l}{F1@5} & \multicolumn{1}{l}{F1@M} & \multicolumn{1}{l}{F1@5} & \multicolumn{1}{l}{F1@M} & \multicolumn{1}{l}{F1@5} & \multicolumn{1}{l}{F1@M} & \multicolumn{1}{l}{F1@5} & \multicolumn{1}{l}{F1@M} \\
 \cline{2-9} 
T $||$ A & 4.68$_1$ & 9.10$_6$ & 6.19$_1$ & 11.89$_2$ & 9.67$_2$ & 18.47$_8$ & 11.97$_1$ & 22.86$_1$ \\
T $||$ A $||$ Body & \inc{26}4.94$_1$ & \inc{23}9.55$_5$ & \dec{20}5.99$_2$ & \dec{14}11.61$_2$ & \inc{12}10.14$_1$ & \inc{27}19.57$_0$ & \inc{17}12.30$_0$ & \inc{17}\textbf{23.53$_0$} \\
\hdashline 
AUG\_TA\_SR & \inc{7}4.75$_1$ & \inc{12}9.34$_3$ & \inc{47}6.66$_2$ & \inc{42}12.74$_0$ & \dec{12}9.19$_3$ & \dec{20}17.65$_{10}$ & \dec{30}11.37$_0$ & \dec{23}21.95$_0$ \\
AUG\_TA\_BT & \dec{27}4.41$_1$ & \dec{24}8.62$_2$ & \inc{13}6.32$_2$ & \inc{19}12.27$_3$ & \inc{19}10.42$_0$ & \inc{37}19.96$_1$ & \inc{19}\textbf{12.34$_0$} & \inc{12}23.32$_2$ \\
AUG\_TA\_KPD & \dec{1}4.67$_1$ & \inc{5}9.19$_1$ & \dec{19}6.00$_0$ & \dec{13}11.63$_1$ & \dec{44}7.92$_2$ & \dec{75}15.48$_5$ & \dec{72}10.53$_0$ & \dec{58}20.55$_1$ \\
AUG\_TA\_KPSR & \dec{13}4.55$_0$ & \dec{8}8.95$_1$ & \dec{49}5.70$_1$ & \dec{49}10.90$_1$ & \dec{63}7.14$_1$ & \dec{115}13.87$_5$ & \dec{132}9.33$_0$ & \dec{114}18.29$_1$ \\
\hdashline
AUG\_Body & \inc{65}\textbf{5.33$_2$} & \inc{66}\textbf{10.42$_5$} & \inc{91}\textbf{7.10$_6$} & \inc{101}\textbf{13.92$_{18}$} & \inc{8}9.97$_5$ & \inc{20}19.25$_{18}$ & \dec{8}11.82$_2$ & \dec{5}22.67$_4$ \\
AUG\_Body\_SR & \inc{20}4.88$_1$ & \inc{29}9.69$_4$ & \inc{31}6.50$_0$ & \inc{32}12.53$_2$ & \dec{8}9.36$_9$ & \dec{8}18.15$_{30}$ & \inc{11}12.19$_1$ & \inc{4}23.04$_3$\\
AUG\_Body\_BT & \dec{9}4.59$_0$ & \dec{3}9.04$_2$ & \inc{17}6.36$_3$ & \inc{18}12.26$_5$ & \inc{21}\textbf{10.50$_0$} & \inc{40}\textbf{20.09$_1$} & \inc{17}12.31$_1$ & \inc{8}23.19$_3$ \\
AUG\_Body\_KPD & \inc{4}4.72$_2$ & \inc{10}9.31$_6$ & \dec{7}6.12$_1$ & \inc{2}11.92$_3$ & \dec{21}8.82$_7$ & \dec{36}17.04$_{18}$ & \dec{18}11.61$_0$ & \dec{18}22.14$_1$ \\
AUG\_Body\_KPSR & \dec{8}4.60$_0$ & \inc{2}9.15$_1$ & \dec{41}5.78$_1$ & \dec{34}11.21$_6$ & \dec{56}7.44$_2$ & \dec{97}14.60$_8$ & \dec{29}11.40$_1$ & \dec{30}21.64$_3$ \\
\hline
\Xhline{3\arrayrulewidth}
 \multirow{2}{*}{\textbf{LDKP10K}} & \multicolumn{2}{c}{1,000} & \multicolumn{2}{c}{2,000} & \multicolumn{2}{c}{4,000} & \multicolumn{2}{c}{8,000} \\
 & \multicolumn{1}{l}{F1@5} & \multicolumn{1}{l}{F1@M} & \multicolumn{1}{l}{F1@5} & \multicolumn{1}{l}{F1@M} & \multicolumn{1}{l}{F1@5} & \multicolumn{1}{l}{F1@M} & \multicolumn{1}{l}{F1@5} & \multicolumn{1}{l}{F1@M} \\
 \cline{2-9}
T $||$ A & \textbf{4.47$_1$} & 8.27$_3$ & 6.66$_1$ & 12.32$_2$ & 9.95$_1$ & 17.49$_2$ & 11.31$_1$ & 19.76$_3$ \\
T $||$ A $||$ Body & \dec{29}3.89$_4$ & \dec{48}7.30$_{14}$ & \dec{6}6.55$_1$ & \dec{6}12.07$_1$ & \dec{14}9.81$_0$ & \inc{3}17.54$_1$ & \inc{39}11.70$_1$ & \inc{24}20.24$_2$ \\
\hdashline
AUG\_TA\_SR & \dec{5}4.37$_0$ & \dec{1}8.26$_0$ & \dec{22}6.22$_0$ & \dec{16}11.67$_1$ & \inc{74}\textbf{10.69$_0$} & \inc{42}\textbf{18.33$_0$} & \inc{99}\textbf{12.30$_0$} & \inc{55}\textbf{20.86$_0$} \\
AUG\_TA\_BT & \dec{22}4.04$_1$ & \dec{34}7.59$_2$ & \inc{52}\textbf{7.70$_1$} & \inc{42}\textbf{14.01$_3$} & \inc{32}10.27$_1$ & \inc{26}18.00$_2$ & \dec{87}10.44$_0$ & \dec{75}18.26$_0$ \\
AUG\_TA\_KPD & \dec{34}3.79$_0$ & \dec{54}7.18$_2$ & \dec{76}5.14$_1$ & \dec{64}9.75$_3$ & \dec{42}9.53$_3$ & \dec{40}16.68$_6$ & \dec{2}11.29$_2$ & \inc{8}19.92$_5$ \\
AUG\_TA\_KPSR & \dec{37}3.74$_0$ & \dec{58}7.11$_1$ & \dec{94}4.77$_1$ & \dec{81}9.08$_3$ & \dec{129}8.66$_3$ & \dec{115}15.19$_6$ & \dec{109}10.22$_0$ & \dec{99}17.78$_1$ \\
\hdashline
AUG\_Body & \dec{1}4.45$_6$ & \inc{9}\textbf{8.45$_{21}$} & \inc{16}6.98$_5$ & \inc{15}12.90$_{12}$ & \inc{42}10.37$_0$ & \inc{40}18.28$_0$ & \inc{61}11.92$_1$ & \inc{48}20.73$_3$ \\
AUG\_Body\_SR & \dec{12}4.22$_0$ & \dec{13}8.01$_0$ & \dec{15}6.36$_0$ & \dec{11}11.88$_1$ & \inc{17}10.12$_0$ & \inc{21}17.91$_0$ & \inc{26}11.57$_0$ & \inc{31}20.38$_1$\\
AUG\_Body\_BT & \dec{4}4.38$_0$ & \dec{5}8.17$_2$ & \inc{50}7.65$_3$ & \inc{39}13.87$_5$ & \inc{29}10.24$_0$ & \inc{23}17.95$_1$ & \dec{14}11.17$_1$ & \dec{2}19.73$_3$\\
AUG\_Body\_KPD & \dec{26}3.96$_3$ & \dec{39}7.49$_8$ & \dec{52}5.61$_2$ & \dec{45}10.54$_4$ & \dec{52}9.43$_0$ & \dec{34}16.81$_0$ & \dec{28}11.03$_0$ & \dec{12}19.53$_0$ \\
AUG\_Body\_KPSR & \dec{32}3.83$_0$ & \dec{51}7.24$_1$ & \dec{95}4.77$_1$ & \dec{82}9.06$_2$ & \dec{71}9.24$_0$ & \dec{50}16.49$_0$ & \dec{38}10.93$_0$ & \dec{25}19.27$_0$ \\
\hline
\Xhline{3\arrayrulewidth}
 \multirow{2}{*}{\textbf{KPTimes}} & \multicolumn{2}{c}{1,000} & \multicolumn{2}{c}{2,000} & \multicolumn{2}{c}{4,000} & \multicolumn{2}{c}{8,000} \\
 & \multicolumn{1}{l}{F1@5} & \multicolumn{1}{l}{F1@M} & \multicolumn{1}{l}{F1@5} & \multicolumn{1}{l}{F1@M} & \multicolumn{1}{l}{F1@5} & \multicolumn{1}{l}{F1@M} & \multicolumn{1}{l}{F1@5} & \multicolumn{1}{l}{F1@M} \\
 \cline{2-9} 
T $||$ A & 9.83$_0$ & 19.01$_1$ & 13.49$_3$ & 24.49$_7$ & 16.92$_1$ & 28.84$_0$ & 19.13$_0$ & 31.89$_0$ \\
T $||$ A $||$ Body & \dec{17}9.66$_2$ & \dec{23}18.56$_6$ & \inc{15}13.64$_2$ & \inc{24}24.98$_4$ & \dec{18}16.74$_0$ & \inc{25}29.33$_0$ & \dec{11}18.91$_0$ & \inc{15}32.20$_0$ \\
\hdashline
AUG\_TA\_SR & \inc{100}11.20$_{10}$ & \inc{100}21.17$_{19}$ & \inc{100}\textbf{15.21$_0$} & \inc{100}26.59$_0$ & \inc{38}\textbf{17.30$_0$} & \inc{26}29.36$_0$ & \inc{9}19.30$_0$ & \inc{27}32.44$_0$ \\
AUG\_TA\_BT & \inc{100}11.02$_4$ & \inc{100}21.22$_8$ & \inc{44}13.93$_3$ & \inc{75}25.99$_7$ & \dec{61}16.31$_2$ & \inc{22}29.29$_3$ & \dec{21}18.71$_1$ & \inc{40}32.69$_0$ \\
AUG\_TA\_KPD & \dec{89}8.94$_0$ & \dec{80}17.41$_0$ & \dec{59}12.90$_1$ & \dec{43}23.63$_3$ & \dec{134}15.58$_1$ & \dec{49}27.86$_0$ & \dec{74}17.64$_1$ & \dec{49}30.91$_1$ \\
AUG\_TA\_KPSR & \dec{71}9.12$_2$ & \dec{57}17.88$_4$ & \inc{34}13.83$_2$ & \inc{20}24.90$_2$ & \dec{115}15.77$_0$ & \dec{62}27.60$_0$ & \dec{57}17.99$_0$ & \dec{48}30.93$_0$ \\
\hdashline
AUG\_Body & \dec{5}9.78$_3$ & \inc{30}19.62$_{11}$ & \inc{82}14.31$_1$ & \inc{88}26.25$_1$ & \inc{34}17.26$_1$ & \inc{74}\textbf{30.33$_1$} & \inc{13}\textbf{19.39$_1$} & \inc{56}33.01$_1$ \\
AUG\_Body\_SR & \inc{100}\textbf{11.21$_4$} & \inc{100}\textbf{22.05$_7$} & \inc{97}14.46$_1$ & \inc{100}\textbf{26.78$_1$} & \dec{6}16.86$_1$ & \inc{65}30.13$_1$ & \dec{9}18.96$_0$ & \inc{67}\textbf{33.23$_0$}\\
AUG\_Body\_BT & \inc{63}10.46$_2$ & \inc{61}20.24$_0$ & \inc{63}14.12$_0$ & \inc{72}25.92$_0$ & \dec{46}16.46$_0$ & \inc{22}29.28$_1$ & \dec{12}18.88$_3$ & \inc{43}32.75$_1$\\
AUG\_Body\_KPD & \dec{103}8.80$_3$ & \dec{69}17.62$_9$ & \dec{13}13.36$_2$ & \inc{13}24.76$_3$ & \dec{43}16.49$_1$ & \inc{30}29.43$_1$ & \dec{32}18.48$_1$ & \inc{14}32.17$_0$ \\
AUG\_Body\_KPSR & \inc{38}10.21$_3$ & \inc{63}20.27$_8$ & \inc{13}13.62$_0$ & \inc{67}25.82$_0$ & \dec{67}16.25$_1$ & \inc{34}29.51$_2$ & \dec{56}18.02$_1$ & \inc{22}32.34$_1$ \\
\hline
\Xhline{3\arrayrulewidth}
\end{tabular}
\caption{Performance for generation of present keyphrases. The results are highlighted with \colorbox{myblue!50}{blue}($\uparrow$) and \colorbox{myred!50}{red}($\downarrow$) with respect to baseline T $||$ A.  $||$ denotes concatenation of the text.
Standard deviation is subscripted to each number and is reported as a multiple of $\pm$ 0.001. Best viewed in color.}
\label{tab:results-present}
\end{table*}

\begin{table*}[ht]
\small
\centering
\renewcommand{\arraystretch}{0.97}
\setlength\tabcolsep{10pt}
\begin{tabular}{lcccccccc}
\hline
\Xhline{3\arrayrulewidth}
\multirow{2}{*}{\textbf{LDKP3K}} & \multicolumn{2}{c}{1,000} & \multicolumn{2}{c}{2,000} & \multicolumn{2}{c}{4,000} & \multicolumn{2}{c}{8,000} \\
& \multicolumn{1}{l}{F1@5} & \multicolumn{1}{l}{F1@M} & \multicolumn{1}{l}{F1@5} & \multicolumn{1}{l}{F1@M} & \multicolumn{1}{l}{F1@5} & \multicolumn{1}{l}{F1@M} & \multicolumn{1}{l}{F1@5} & \multicolumn{1}{l}{F1@M} \\
 \cline{2-9}  
T $||$ A & 0.078$_0$ & 0.169$_0$ & 0.129$_0$ & 0.281$_0$ & 0.044$_0$ & 0.093$_0$ & 0.044$_0$ & 0.099$_0$ \\
T $||$ A $||$ Body & \inc{1}0.079$_0$ & \dec{2}0.165$_0$ & \inc{1}0.130$_0$ & \inc{0}0.282$_0$ & \inc{2}0.047$_0$ & \inc{3}0.105$_0$ & \dec{6}0.031$_0$ & \dec{6}0.073$_0$ \\
\hdashline
AUG\_TA\_SR & \inc{54}0.132$_0$ & \inc{60}0.290$_0$ & \inc{7}0.136$_0$ & \inc{10}0.300$_0$ & \inc{39}0.096$_0$ & \inc{29}0.207$_0$ & \inc{12}0.067$_0$ & \inc{10}0.141$_0$ \\
AUG\_TA\_BT & \inc{50}0.128$_0$ & \inc{55}0.279$_0$ & \inc{10}0.139$_0$ & \inc{12}0.305$_0$ & \inc{18}0.068$_0$ & \inc{12}0.140$_0$ & \inc{39}0.121$_0$ & \inc{42}0.266$_0$ \\
AUG\_TA\_KPD & \inc{62}0.140$_0$ & \inc{71}0.311$_0$ & \inc{16}0.145$_0$ & \inc{18}0.318$_0$ & \inc{72}0.141$_0$ & \inc{54}0.307$_0$ & \inc{28}0.099$_0$ & \inc{30}0.218$_0$ \\
AUG\_TA\_KPSR & \inc{64}0.142$_0$ & \inc{69}0.307$_0$ & \inc{48}0.177$_0$ & \inc{56}0.393$_0$ & \inc{80}0.151$_0$ & \inc{57}0.321$_0$ & \inc{55}0.154$_0$ & \inc{57}0.325$_0$ \\
\hdashline
AUG\_Body & \inc{51}0.129$_0$ & \inc{61}0.291$_0$ & \inc{1}0.130$_0$ & \inc{6}0.292$_0$ & \inc{13}0.061$_0$ & \inc{11}0.138$_0$ & \inc{18}0.079$_0$ & \inc{19}0.175$_0$ \\
AUG\_Body\_SR & \inc{63}0.141$_0$ & \inc{75}0.319$_0$ & \inc{28}0.157$_0$ & \inc{30}0.342$_0$ & \inc{24}0.076$_0$ & \inc{17}0.161$_0$ & \inc{52}0.149$_0$ & \inc{56}0.322$_0$\\
AUG\_Body\_BT & \inc{52}0.130$_0$ & \inc{59}0.287$_0$ & \dec{8}0.121$_0$ & \dec{8}0.265$_0$ & \inc{28}0.081$_0$ & \inc{23}0.183$_0$ & \inc{38}0.120$_0$ & \inc{38}0.253$_0$ \\
AUG\_Body\_KPD & \inc{66}0.144$_0$ & \inc{79}0.328$_0$ & \inc{60}0.189$_0$ & \inc{63}0.407$_0$ & \inc{69}0.136$_0$ & \inc{51}0.298$_0$ & \inc{69}0.182$_0$ & \inc{75}0.398$_0$ \\
AUG\_Body\_KPSR & \inc{84}\textbf{0.162$_0$} & \inc{95}\textbf{0.359$_0$} & \inc{71}\textbf{0.200$_0$} & \inc{80}\textbf{0.441$_0$} & \inc{100}\textbf{0.184$_0$} & \inc{78}\textbf{0.405$_0$} & \inc{92}\textbf{0.227$_0$} & \inc{99}\textbf{0.495$_0$} \\
\hline
\Xhline{3\arrayrulewidth}
 \multirow{2}{*}{\textbf{LDKP10K}} & \multicolumn{2}{c}{1,000} & \multicolumn{2}{c}{2,000} & \multicolumn{2}{c}{4,000} & \multicolumn{2}{c}{8,000} \\
 & \multicolumn{1}{l}{F1@5} & \multicolumn{1}{l}{F1@M} & \multicolumn{1}{l}{F1@5} & \multicolumn{1}{l}{F1@M} & \multicolumn{1}{l}{F1@5} & \multicolumn{1}{l}{F1@M} & \multicolumn{1}{l}{F1@5} & \multicolumn{1}{l}{F1@M} \\
 \cline{2-9} 
T $||$ A & 0.023$_0$ & 0.047$_0$ & 0.039$_0$ & 0.079$_0$ & 0.114$_0$ & 0.228$_0$ & 0.184$_0$ & 0.335$_0$ \\
T $||$ A $||$ Body & \dec{6}0.021$_0$ & \dec{7}0.044$_0$ & \dec{4}0.035$_0$ & \dec{2}0.074$_0$ & \dec{31}0.052$_0$ & \dec{31}0.105$_0$ & \dec{6}0.159$_0$ & \dec{9}0.289$_0$ \\
\hdashline
AUG\_TA\_SR & \inc{32}0.031$_0$ & \inc{27}0.061$_0$ & \inc{15}0.054$_0$ & \inc{15}0.110$_0$ & \inc{41}0.195$_0$ & \inc{40}0.387$_0$ & \inc{43}0.355$_0$ & \inc{59}0.629$_0$ \\
AUG\_TA\_BT & \inc{15}0.027$_0$ & \inc{7}0.051$_0$ & \inc{45}0.084$_0$ & \inc{47}0.173$_0$ & \inc{41}0.196$_0$ & \inc{39}0.383$_0$ & \inc{38}0.337$_0$ & \inc{56}0.617$_0$ \\
AUG\_TA\_KPD & \dec{12}0.020$_0$ & \dec{12}0.041$_0$ & \inc{18}0.057$_0$ & \inc{18}0.115$_0$ & \inc{48}0.210$_0$ & \inc{44}0.403$_0$ & \inc{29}0.299$_0$ & \inc{43}0.552$_0$ \\
AUG\_TA\_KPSR & \inc{32}0.031$_0$ & \inc{25}0.059$_0$ & \inc{28}0.067$_0$ & \inc{27}0.133$_0$ & \inc{57}0.229$_0$ & \inc{51}0.433$_0$ & \inc{61}0.429$_0$ & \inc{87}0.769$_0$ \\
\hdashline
AUG\_Body & \inc{38}0.033$_0$ & \inc{32}0.063$_0$ & \inc{32}0.071$_0$ & \inc{35}0.148$_0$ & \inc{46}0.206$_0$ & \inc{45}0.407$_0$ & \inc{40}0.344$_0$ & \inc{57}0.622$_0$ \\
AUG\_Body\_SR & \inc{57}0.037$_0$ & \inc{48}0.071$_0$ & \inc{46}0.085$_0$ & \inc{44}0.168$_0$ & \inc{49}0.213$_0$ & \inc{45}0.410$_0$ & \inc{48}0.378$_0$ & \inc{70}0.687$_0$\\
AUG\_Body\_BT & \inc{41}0.033$_0$ & \inc{33}0.064$_0$ & \inc{34}0.073$_0$ & \inc{36}0.151$_0$ & \inc{39}0.193$_0$ & \inc{40}0.387$_0$ & \inc{38}0.338$_0$ & \inc{60}0.637$_0$ \\
AUG\_Body\_KPD & \inc{86}0.044$_0$ & \inc{81}0.088$_0$ & \inc{46}0.085$_0$ & \inc{44}0.166$_0$ & \inc{62}0.238$_0$ & \inc{59}0.465$_0$ & \inc{54}0.400$_0$ & \inc{78}0.726$_0$ \\
AUG\_Body\_KPSR & \inc{89}\textbf{0.045$_0$} & \inc{83}\textbf{0.089$_0$} & \inc{67}\textbf{0.106$_0$} & \inc{65}\textbf{0.210$_0$} & \inc{73}\textbf{0.259$_0$} & \inc{66}\textbf{0.492$_0$} & \inc{69}\textbf{0.459$_0$} & \inc{98}\textbf{0.827$_0$} \\
\hline
\Xhline{3\arrayrulewidth}
 \multirow{2}{*}{\textbf{KPTimes}} & \multicolumn{2}{c}{1,000} & \multicolumn{2}{c}{2,000} & \multicolumn{2}{c}{4,000} & \multicolumn{2}{c}{8,000} \\
 & \multicolumn{1}{l}{F1@5} & \multicolumn{1}{l}{F1@M} & \multicolumn{1}{l}{F1@5} & \multicolumn{1}{l}{F1@M} & \multicolumn{1}{l}{F1@5} & \multicolumn{1}{l}{F1@M} & \multicolumn{1}{l}{F1@5} & \multicolumn{1}{l}{F1@M} \\
 \cline{2-9} 
T $||$ A & 0.026$_0$ & 0.051$_0$ & 0.026$_0$ & 0.247$_0$ & 1.430$_0$ & 2.445$_1$ & 3.066$_0$ & 5.393$_1$ \\
T $||$ A $||$ Body & \dec{3}0.023$_0$ & \dec{3}0.044$_0$ & \dec{3}0.023$_0$ & \inc{1}0.271$_0$ & \dec{17}1.082$_0$ & \dec{12}1.950$_0$ & \dec{25}2.558$_0$ & \dec{17}4.719$_0$ \\
\hdashline
AUG\_TA\_SR & \inc{79}0.105$_0$ & \inc{63}0.176$_0$ & \inc{100}1.240$_0$ & \inc{96}2.168$_0$ & \inc{64}2.718$_0$ & \inc{55}4.648$_0$ & \inc{60}4.274$_0$ & \inc{49}7.336$_0$ \\
AUG\_TA\_BT & \inc{100}0.163$_0$ & \inc{100}0.277$_0$ & \inc{100}0.163$_1$ & \inc{90}2.050$_2$ & \inc{54}2.501$_0$ & \inc{49}4.390$_0$ & \inc{38}3.826$_1$ & \inc{36}6.818$_1$ \\
AUG\_TA\_KPD & \inc{34}0.060$_0$ & \inc{28}0.107$_0$ & \inc{34}0.060$_0$ & \inc{20}0.637$_0$ & \inc{10}1.634$_0$ & \inc{10}2.854$_0$ & \inc{18}3.423$_0$ & \inc{15}6.006$_0$ \\
AUG\_TA\_KPSR & \inc{61}0.087$_0$ & \inc{56}0.163$_0$ & \inc{61}0.087$_0$ & \inc{80}1.841$_1$ & \inc{66}\textbf{2.748$_0$} & \inc{55}4.648$_0$ & \inc{70}\textbf{4.465$_0$} & \inc{49}7.352$_0$ \\
\hdashline
AUG\_Body & \inc{7}0.033$_0$ & \inc{4}0.060$_0$ & \inc{7}0.033$_0$ & \inc{45}1.150$_0$ & \inc{52}2.460$_0$ & \inc{48}4.380$_0$ & \inc{55}4.171$_0$ & \inc{50}7.385$_0$ \\
AUG\_Body\_SR & \inc{100}0.159$_0$ & \inc{100}0.278$_0$ & \inc{100}1.122$_0$ & \inc{88}1.999$_0$ & \inc{63}2.681$_0$ & \inc{57}4.708$_0$ & \inc{53}4.135$_1$ & \inc{47}7.285$_1$\\
AUG\_Body\_BT & \inc{100}0.131$_0$ & \inc{94}0.239$_0$ & \inc{100}\textbf{1.191$_0$} & \inc{95}\textbf{2.152$_0$} & \inc{47}2.363$_0$ & \inc{44}4.215$_1$ & \inc{36}3.795$_0$ & \inc{32}6.670$_1$ \\
AUG\_Body\_KPD & \inc{12}0.038$_0$ & \inc{9}0.069$_0$ & \inc{12}0.038$_0$ & \inc{48}1.200$_1$ & \inc{58}2.588$_0$ & \inc{54}4.607$_0$ & \inc{66}4.382$_0$ & \inc{55}7.575$_0$ \\
AUG\_Body\_KPSR & \inc{100}\textbf{0.182$_0$} & \inc{100}\textbf{0.319$_0$} & \inc{100}0.182$_0$ & \inc{86}1.963$_1$ & \inc{64}2.708$_0$ & \inc{57}\textbf{4.744$_0$} & \inc{63}4.325$_1$ & \inc{56}\textbf{7.629$_2$} \\
\hline
\Xhline{3\arrayrulewidth}
\end{tabular}
\caption{Performance for generation of absent keyphrases. The results are highlighted with \colorbox{myblue!50}{blue}($\uparrow$) and \colorbox{myred!50}{red}($\downarrow$) with respect to baseline T $||$ A.  $||$ denotes concatenation of the text.
Standard deviation is subscripted to each number and is reported as a multiple of $\pm$ 0.001. Best viewed in color.}
\label{tab:results-absent}
\end{table*}

\subsection{Evaluation}
We compare the performance of the different methods comprehensively for four low-resource settings, i.e., with 1000, 2000, 4000 and 8000 samples. The settings are highly competitive to the prior works where they used at best 5000 samples \cite{chowdhury2021kpdrop, wu-etal-2022-representation} for their experiments. Following prior works \cite{meng-etal-2017-deep, chen-etal-2018-keyphrase, chan-etal-2019-neural,chen-etal-2020-exclusive}, we report the results for metrics F1@5\footnote{We use the metrics from \cite{chan-etal-2019-neural} and adopted by \citet{chen-etal-2020-exclusive, ahmad-etal-2021-select, ye-etal-2021-heterogeneous}.} and F1@M in the main tables. All comparisons are done after stemming the text as well as keyphrases.

Following \citet{meng-etal-2017-deep, chan-etal-2019-neural, yuan-etal-2020-one}, we use GRU encoder-decoder-based architecture for evaluating all models. For all experiments, we restrict the length of the body (or equivalently, full text) to a maximum sequence length of 800 words. For each setting, we sample thrice and further repeat each sample for three different seeds. We thus report the average result for a total of nine runs (3 samples * 3 seeds) for each setting. Hyperparameters and other implementation details are presented in Appendix \S \ref{sec:implementation}.

\begin{table*}[t]
\begin{small}
\centering
\resizebox{\textwidth}{!}{
\setlength\tabcolsep{10pt}
\begin{tabular}{p{0.65\linewidth} p{0.10\linewidth} p{0.25\linewidth}}
\toprule
\multicolumn{1}{c}{\rule{0pt}{2ex}\textbf{Excerpts from test dataset samples}} &  \multicolumn{1}{c}{\textbf{Methods}} & \multicolumn{1}{c}{\textbf{Predicted Keyphrases}}\\
\hline
\rule{0pt}{3ex} \makecell{\multirow{4}{10cm}{committees of learning agents [SEP] we describe how machine learning and decision theory is combined in an application that supports control room operators of a combined heating and power plant ... \\ \textbf{Gold:} \colorbox{myblue!50}{machine learning}; \colorbox{myblue!50}{committees}; \colorbox{myred!50}{decision analysis}}} & T $||$ A & learning\\[0.5ex] 
& Aug\_Body & \colorbox{myblue!50}{machine learning} \\[1ex]
& Aug\_Body\_SR & learning\\[1ex] 
\hline
\rule{0pt}{3ex} \makecell{\multirow{4}{10cm}{compositional analysis for linear control systems [SEP] the complexity of physical and engineering systems , both in terms of the governing physical phenomena and the number of subprocesses involved ...
\\\textbf{Gold:} \colorbox{myred!50}{compositional reasoning}; \colorbox{myred!50}{linear systems}; \colorbox{myblue!50}{simulation relations}; \colorbox{myred!50}{assume-guarantee reasoning}}} & T $||$ A & control\\[2.5ex] 
& Aug\_Body & linear control; \colorbox{myred!50}{linear systems}\\[2.5ex]
& Aug\_Body\_SR & linear control; \colorbox{myred!50}{linear systems}\\[1ex] 
\hline
\rule{0pt}{3ex} \makecell{\multirow{4}{10cm}{the bits and flops of the n-hop multilateration primitive for node localization problems [SEP] the recent advances in mems , embedded systems and wireless communication technologies are making the realization ...\\ \textbf{Gold:} \colorbox{myblue!50}{technologies}; \colorbox{myred!50}{ad-hoc localization}; \colorbox{myred!50}{sensor networks}; \colorbox{myblue!50}{embedded systems}; \colorbox{myblue!50}{wireless}; \colorbox{myblue!50}{network}}} & T $||$ A & tangible\\[2.5ex]
& Aug\_Body & wireless networks\\[2.5ex]
& Aug\_Body\_SR & sensors\\[1.4ex]
\bottomrule
\end{tabular}
}
\caption{Sample predictions using models trained with different (representative) augmentation methods and the baseline (T $||$ A). The text is highlighted as follows: \colorbox{myblue!50}{\textsc{Present Keyphrases}}, \colorbox{myred!50}{\textsc{Absent Keyphrases}}. Note that the test samples contain only T || A. Best viewed in color.}
\label{tab:predictions}
\end{small}
\end{table*}

\section{Results and Analysis}
\label{sec:results}
We present our discussion of results for the generation of the two types of keyphrases, i.e., \textit{present} and \textit{absent} in \S\ref{sec:results-present} and \S\ref{sec:results-absent}, respectively.

\subsection{Present Keyphrase Generation}
\label{sec:results-present}
From Table \ref{tab:results-present}, we make the following observations. First, augmenting the baseline T $||$ A with the text from the body (\textsc{Aug\_Body}) helps to improve the present keyphrase generation performance. Second, we observe that the methods that use the body (prepended with \textsc{Aug\_Body}) are better than the augmentation methods that just use Title and Abstract (prepended with \textsc{Aug\_TA}). These two observations imply that the body constitutes a rich source of present keyphrases. 

Third, we also compare with \citet{garg-etal-2022-keyphrase} (\textsc{T $||$ A $||$ Body}) where they concatenated different types of sentences to T $||$ A. 
We observe that augmenting the text from the articles (\textsc{Aug\_Body}) instead of merely concatenating them (\textsc{T $||$ A $||$ Body}) improves the performance by a wide margin. It is also interesting to observe that \textsc{T $||$ A $||$ Body}, which found significant performance gains in large-scale settings, underperforms even T $||$ A in many purely low-resource settings.

Fourth, the results suggest a quite intriguing observation that the standard data augmentation techniques like synonym replacement and back translation (suffixed with \textsc{SR, BT}) are more rewarding for present keyphrase generation performance than the techniques specifically designed for the keyphrase generation task (suffixed with \textsc{KPD, KPSR}). This trend could be because synonym replacement and back translation bring more diversity to the training samples (since they replace/ rephrase a much larger portion of the text) compared to keyphrase-specific techniques which modify only a handful of tokens (i.e., present keyphrases) in the text. It is worth mentioning that even these standard data augmentation techniques have been largely ignored by the current research on keyphrase generation. 

Fifth, we rather observe that the keyphrase-specific data augmentation techniques are not just lower in performance than the standard data augmentation techniques but often they hurt the performance of the model when trained in purely low-resource settings. The reason could be that the models do not have enough samples and diversity to learn to generate the present keyphrases, all the more when the present keyphrases are dropped or replaced during training. This is in contrast with the behavior of models when trained on a large-scale dataset, where the performance of present keyphrase generation (\textsc{Aug\_TA\_KPD}) is on par with T $||$ A \cite{chowdhury2021kpdrop}. 

Sixth, in Table \ref{tab:results-present}, we can also compare the performance of models trained on: (1) total $x$ original samples, (2) $x$ original + $x$ augmented samples, (3) total $2x$ original samples. For example, for LDKP3K dataset, we observe that 2000 original samples achieve the best performance (11.89 in F1@M), followed by the augmented version (9.34 for augmentation with synonym replacement, 10.42 for augmentation with body) whereas the performance when using 1000 original samples is 9.10. We observe similar trends across the different augmentation strategies and datasets.

We draw the following conclusions: (1) Data augmentation techniques for keyphrase generation have been quite an under-studied topic, particularly for low-resource settings and the behavior of the models is different than that when training on large-scale settings; (2) We show that existing works such as those by \citet{garg-etal-2022-keyphrase, chowdhury2021kpdrop} can be surpassed by the data augmentation methods discussed in this work when used in low-resource settings for {\em present} keyphrase generation. 

\subsection{Absent Keyphrase Generation}
\label{sec:results-absent}
To investigate the ability of the KG models to develop a semantic understanding of the documents, we evaluate the performance of the absent keyphrase generation. Table \ref{tab:results-absent} presents the absent keyphrase performance of the different augmentation methods. Our observations are as follows. First, augmentation with the body (prefixed with \textsc{Aug\_Body}) still surpasses the Title and Abstract (prefixed with \textsc{Aug\_TA}) counterparts. Second, unlike the present keyphrase generation performance, the absent keyphrase generation performance is generally better with almost all the data augmentation methods compared to the baseline T $||$ A. The reason could be that the augmentation methods artificially turn some of the present keyphrases to absent keyphrases (e.g., present keyphrases replaced with synonyms or dropped or rephrased). Thus, the model finds much more opportunities to learn to generate absent keyphrases.

Third, interestingly, KG-targeted data augmentation methods (suffixed with \textsc{KPD, KPSR}) perform better than the standard data augmentation methods like synonym replacement and back translation (suffixed with \textsc{SR, BT}) for generating absent keyphrases (unlike present keyphrase generation). This is because \textsc{KPD, KPSR} specifically replace the present keyphrases to become absent keyphrases. Whereas \textsc{SR, BT} \textit{randomly} replace/ rephrase the tokens and thus, one would expect a less number of present keyphrases turning into absent keyphrases. Fourth, augmentation with KG-based synonym replacement (\textsc{KPSR}) surpasses even the dropout augmentation technique (\textsc{KPD}). This might be because of two reasons: (1) the keyphrase dropout method masks the keyphrases with some probability value whereas we replace all the present keyphrases with their synonyms, (2) dropping the important keyphrases hides some information from the model, while replacing the keyphrases with their synonyms still largely preserves the semantics and integrity of the text.

Fifth, we observe that the model proposed by \citet{garg-etal-2022-keyphrase} which is based on concatenation is not able to generalize well in the low-resource settings, rather, ends up weakening the model performance compared to T $||$ A. This again urges towards the development of data augmentation methods in purely low-data regimes.

Sixth, in Table \ref{tab:results-absent},
the results show that the model trained on the combination of original and augmented samples outperforms the settings where the model is trained on equivalent amount of original samples, for most datasets and augmentation strategies. For instance, for LDKP3K dataset, the 2000 augmentation version achieves 0.290 in F1@M (for augmentation with synonym replacement on Title and Abstract) and outperforms both 2000 original samples (0.281) and 1000 original samples (0.169). Thus, for the same amount of data (2000 dataset size), the augmented version shows better results than without data augmentation.

We show sample predictions from the representative models: \textsc{T $||$ A} (baseline), \textsc{Aug\_Body} (best for Present KG), \textsc{Aug\_Body\_SR} (best for Absent KG) in Table \ref{tab:predictions}. In the table, we can observe that while T $||$ A fails to capture the specific topics (or keyphrases) for the document, models trained with augmentation strategies can generalize better.

\section{Analysis}
\label{sec:case-study}
In this section, we study one of the settings in more detail, i.e., with the LDKP3K dataset having 1000 samples in the training set (and twice the number in the training set for \textsc{Aug}-prefixed methods). The study unfolds into two aspects: (a) analyzing the data created for the different augmentation methods, (b) developing better inference strategies.

\begin{table}[t]
\small
\centering
\renewcommand{\arraystretch}{1.1}
\setlength\tabcolsep{8pt}
\begin{tabular}{lccc}
\hline
\Xhline{3\arrayrulewidth}
 \multicolumn{1}{l}{Methods} & \multicolumn{1}{l}{Pres.KP} & \multicolumn{1}{l}{Abs.KP} & \multicolumn{1}{l}{TotalKP} \\
 \hline
T $||$ A & 3374 & 2093 & 5467 \\
T $||$ A $||$ Body & 3985 & 1482 & 5467 \\
AUG\_TA\_SR & 5761 & 5173 & 10934 \\
AUG\_TA\_BT & 5499 & 5435 & 10934 \\
AUG\_TA\_KPD & 4586 & 6348 & 10934 \\
AUG\_TA\_KPSR & 4532 & 6402 & 10934 \\
AUG\_Body & \textbf{6309} & 4625 & 10934 \\
AUG\_Body\_SR & 5402 & 5532 & 10934 \\
AUG\_Body\_BT & 5291 & 5643 & 10934 \\
AUG\_Body\_KPD & 4590 & \textbf{6344} & 10934 \\
AUG\_Body\_KPSR & 4591 & 6343 & 10934 \\
\Xhline{3\arrayrulewidth}
\end{tabular}
\renewcommand{\arraystretch}{1}
\setlength\tabcolsep{2pt}
\begin{tabular}{lllll}
\hline
\Xhline{3\arrayrulewidth}
\end{tabular}
\caption{{Number of present (Pres.KP), absent (Abs.KP), total keyphrases (TotalKP) in the training set of LDKP3K with 1000 samples for the different augmentation methods. }
\label{tab:analysis}}
\end{table}

We analyze the data created using the different augmentation methods and report the present, absent and total number of keyphrases in Table \ref{tab:analysis}. First, we observe that all the data augmentation methods have double the total number of keyphrases because the total number of samples is doubled. In effect, the model develops a better generalization ability when it practices with more instances of present and absent keyphrases. Second, we see that \textsc{Aug\_Body} has the highest number of present keyphrases. This implies that the text from the body of the articles not only adds diversity to the training samples (as also evident from Tables \ref{tab:cherryexample}, \ref{tab:results-present}), but also the diversity contains a lot of present keyphrases, unlike other augmentation methods like \textsc{KPD, KPSR}. Third, it is also evident from Table \ref{tab:analysis} that the KG-specific data augmentation methods (suffixed with \textsc{KPD, KPSR}) are rich sources of absent keyphrases whereas the standard data augmentation (suffixed with \textsc{SR, BT}) methods are rich in present keyphrases. This further explains the observations made in the previous sections \S\ref{sec:results-present}-\ref{sec:results-absent} that the KG-specific augmentation methods perform better for absent keyphrase generation, whereas the standard data augmentation methods do better in present keyphrase generation.

Further, in Table \ref{tab:case-study}, we present some of the representative inference strategies by unionizing different augmentation methods during inference. \textit{Union} can be seen as a post-training augmentation method that (during inference) takes a union of the predictions from multiple models that are pretrained using different augmentation methods. The idea is to leverage the complementary strength of the different models that are good for either or both present and absent keyphrase generation. As expected, the performance of the \textit{Union} methods surpasses that of the individual augmentation methods. 


\begin{table}[t]
\small
\centering
\renewcommand{\arraystretch}{1.1}
\setlength\tabcolsep{8pt}
\begin{tabular}{lccc}
\hline
\Xhline{3\arrayrulewidth}
\end{tabular}
\renewcommand{\arraystretch}{1}
\setlength\tabcolsep{2pt}
\begin{tabular}{lllll}
\hline
\Xhline{3\arrayrulewidth}
 \hline
\multirow{2}{*}{Methods}  & \multicolumn{2}{c}{Present} & \multicolumn{2}{c}{Absent} \\
\multicolumn{1}{l}{} & F1@5 & F1@M & F1@5 & F1@M \\
\hline
T $||$ A & 4.68 & 9.10 & 0.078 & 0.169 \\
AUG\_TA\_BT & 4.41 & 8.62 & 0.128 & 0.279 \\
AUG\_TA\_KPSR & 4.55 & 8.95 & 0.132 & 0.290 \\
AUG\_Body & \textbf{5.33} & \textbf{10.42} & 0.129 & 0.291 \\
AUG\_Body\_BT & 4.59 & 9.04 & 0.130 & 0.287 \\
AUG\_Body\_KPD & 4.72 & 9.31 & 0.144 & 0.328 \\
AUG\_Body\_KPSR & 4.60 & 9.15 & \textbf{0.162} & \textbf{0.359} \\
\hdashline
\rule{0pt}{3ex}\textbf{Inference Strategies} & & & & \\
\rule{0pt}{3ex}
Body $\cup$ Body-KPSR & 6.41 & \textbf{11.95} & 0.196 & 0.428 \\
TA-BT $\cup$ Body-BT & 5.39 & 10.19 & 0.160 & 0.342\\
TA-KPSR $\cup$ Body-KPSR & 6.17 & 11.47 & \textbf{0.220} & \textbf{0.462} \\
Body-BT $\cup$ Body-KPD & \textbf{6.45} & 11.81 & 0.204 & 0.435 \\
Body-KPSR $\cup$ Body-KPD & 5.94 & 11.18 & 0.204 & 0.444 \\
\hline
\Xhline{3\arrayrulewidth}
\end{tabular}
\caption{A comparison of various Inference Strategies using \textit{Union} (see \S\ref{sec:case-study}) with the individual (AUG$\_$) methods on LDKP3K with 1000 samples in the training set.}
\label{tab:case-study}
\vspace{-1.4em}
\end{table}

\section{Conclusion}
Although data augmentation has been a very common practice to advance the state-of-the-art in NLP, it has been under-explored for the keyphrase generation (KG) task. Thus, this work discusses various data augmentation methods including both types (i.e., standard and KG-specific) particularly for purely low-resource keyphrase generation, and provides comprehensive evaluation for 12 different settings (four settings for three datasets each).

We also leverage the full text of the articles for data augmentation and observe large improvements over the baseline as well as over data augmentation methods that use only title and abstract (T $||$ A). Detailed analysis helps us believe that KG-specific data augmentation methods can largely improve absent keyphrase generation but at the cost of present keyphrase generation. In contrast, the standard data augmentation techniques like synonym replacement and back-translation are capable of introducing enough diversity to improve the present keyphrase generation without bringing a drop in absent keyphrase generation performance. Although augmentation with the body improves both types of generation to some degree, this work leaves much room to develop better data augmentation strategies to train the model to do better on both present and absent keyphrase generation in low-resource settings which are prevalent in many domains.

\section{Limitations}                       
\label{sec:limitations}
We conducted extensive experiments with three datasets from different domains to substantiate the results thoroughly. 
We observe the best performance when we also leverage the body of the articles. So, we did not evaluate the performance on the datasets that do not have the full text (or equivalently, long text) of the articles. 


\section{Ethics Statement}
In our work, we provide a comprehensive analysis and present data augmentation strategies specifically to address keyphrase generation in purely resource-constrained domains. We do not expect any direct ethical concern from our work.

\section*{Acknowledgments}
This research is supported in part by NSF CAREER
award \#1802358, NSF CRI award \#1823292, NSF
IIS award \#2107518, and UIC Discovery Partners
Institute (DPI) award. Any opinions, findings, and
conclusions expressed here are those of the authors
and do not necessarily reflect the views of NSF or
DPI. We thank AWS for computational resources
used for this study. We also thank our anonymous
reviewers for their valuable and constructive feedback and suggestions.

\bibliography{anthology,custom}
\bibliographystyle{acl_natbib}

\newpage

\appendix
\section{More Implementation Details}
\label{sec:implementation}

Following \citet{garg-etal-2022-keyphrase}, we preprocessed the full text of the articles for all three datasets. We filtered all the articles that had either of the four fields missing, viz., title, abstract, keyphrases, full text, or that contained less than five sentences in the full text. We segmented the full text into sentences using PunktSentenceTokenizer\footnote{\url{https://www.nltk.org/_modules/nltk/tokenize/punkt.html}} and tokenized the sentences further into tokens using NLTK's word\_tokenizer. We also lowercased the text, removed html text, emails, urls, escape symbols, and converted all the numbers into <digit> \cite{meng-etal-2017-deep}, and finally removed any duplicate items in the collection. Further, we subsampled the datasets to construct four low-resource settings (sampled thrice for each setting) containing 1000, 2000, 4000 and 8000 samples.

We use the GRU-based architecture for evaluating all the methods. Similar to \citet{meng-etal-2017-deep, yuan-etal-2020-one, chan-etal-2019-neural} we use an encoder-decoder architecture (where both the encoder and the decoder are GRUs) with attention and a pointer mechanism \cite{see-etal-2017-get}. The exact details of the architecture are similar to that of \citet{chan-etal-2019-neural}. The vocabulary size is 50,000 and each word is translated into embeddings of dimension equal to 100. The GRU encoders and decoders have hidden layer sizes of 150 and 300 respectively. We use a learning rate of 1e-3, batch size of 4, Adam optimizer, ReduceLROnPlateau scheduler and maximum epochs as 20. We early stop the training with patience value of 2. 

\end{document}